\documentclass{article}

\usepackage{microtype}
\usepackage{graphicx}
\usepackage{subcaption}
\usepackage{booktabs} 

\usepackage{hyperref}

\makeatletter
\g@addto@macro\UrlBreaks{%
  \do\a\do\b\do\c\do\d\do\e\do\f\do\g\do\h\do\i\do\j
  \do\k\do\l\do\m\do\n\do\o\do\p\do\q\do\r\do\s\do\t
  \do\u\do\v\do\w\do\x\do\y\do\z
  \do\A\do\B\do\C\do\D\do\E\do\F\do\G\do\H\do\I\do\J
  \do\K\do\L\do\M\do\N\do\O\do\P\do\Q\do\R\do\S\do\T
  \do\U\do\V\do\W\do\X\do\Y\do\Z
  \do\0\do\1\do\2\do\3\do\4\do\5\do\6\do\7\do\8\do\9}
\makeatother
\newcommand{\deq}{\overset{\text{def}}{=}}

\usepackage[preprint]{icml2026/icml2026}

\usepackage{amsmath}
\usepackage{amssymb}
\usepackage{mathtools}
\usepackage{amsthm}

\usepackage[capitalize,noabbrev]{cleveref}

\theoremstyle{plain}

\theoremstyle{definition}

\theoremstyle{remark}

\usepackage[disable,textsize=tiny]{todonotes}

\icmltitlerunning{Backtracking Bursts in Long Reasoning Traces}

\begin{document}

\twocolumn[
  \icmltitle{The Shape of Overthinking: Backtracking Bursts in Long Reasoning Traces}

  \begin{icmlauthorlist}
    \icmlauthor{Navid Rezazadeh}{uci}
    \icmlauthor{Arash Gholami Davoodi}{cmu}
  \end{icmlauthorlist}

  \icmlaffiliation{uci}{University of California, Irvine}
  \icmlaffiliation{cmu}{Carnegie Mellon University}

  \icmlcorrespondingauthor{Navid Rezazadeh}{nrezazad@uci.edu}
  \icmlcorrespondingauthor{Arash Gholami Davoodi}{agholami@andrew.cmu.edu}

  \icmlkeywords{large language models, mathematical reasoning, test-time compute, early exit, efficient inference}

  \vskip 0.3in
]

\printAffiliationsAndNotice{}

\begin{abstract}
Reasoning models often generate long traces in which useful self-correction and unproductive revision are hard to distinguish. We study this distinction through backtracking dynamics: local reconsideration, retraction, or re-derivation inside long-form reasoning traces. On 6{,}000 Qwen3-8B AIME traces, we annotate segment-level backtrack severity and analyze event timing, normalized depth, and local burst structure. We find that early isolated repair is often compatible with correct reasoning, whereas incorrect traces more often show moderate-to-severe backtracks that persist and cluster late. Cross-corpus checks show the same qualitative asymmetry across additional model/domain pairs. Filtering analyses instantiate the signal as a prefix-causal selective early-exit policy: at shallow and intermediate depths, burst-aware filtering outperforms fixed length-based filtering while using only prefix-available features. Moderate length cutoffs remain strong completed-trace baselines, but burst-aware control provides a deployable mechanism for separating recoverable repair from likely instability.
\end{abstract}

\section{Introduction}
\label{sec:introduction}

Reasoning-oriented language models often improve when given more test-time compute, but additional reasoning is not uniformly helpful. Some traces use extra computation to repair a local mistake and then proceed without further revision; others continue revising intermediate states long after useful progress has slowed. This makes adaptive stopping appealing, but also difficult: a useful signal must distinguish productive self-correction from unproductive persistence \cite{graves2016adaptive,teerapittayanon2016branchynet,delcorro2023skipdecode,elhoushi2024layerskip,liu2025answerconvergence,wei2025stopspinningwheels,sharma2025thinkjustenough}.

We study this distinction through \emph{backtracking dynamics}. By \emph{backtrack} we mean a local act of reconsideration inside a reasoning trace: retracting a tentative conclusion, revisiting an assumption, redoing a derivation, or abandoning one line of attack for another. A little self-correction is part of healthy reasoning; what appears more concerning is a different regime in which moderate-to-severe backtracks recur, persist, and begin to cluster later in the trace---a local clustering we call \emph{burst structure}. The key question is therefore not only \emph{how much} the model thinks, but \emph{how} revision is organized over time: early isolated repair is often compatible with success, whereas late clustered reversal is a more plausible marker of diminishing returns.

To study this, we use 6{,}000 reasoning traces from 60 AIME problems generated by Qwen3-8B with explicit thinking enabled as the primary corpus. We segment each trace into scored steps, assign move labels, and attach a backtrack-confidence score to each segment. This lets us compare correct and incorrect traces along three complementary axes: when qualifying backtracks first appear, how often they occur at matched relative progress, and whether they remain isolated or collapse into bursts. We then repeat the same signature checks on additional model and domain pairs to test whether the pattern is specific to this primary setting.

This paper makes three contributions: (i) a trace-level characterization of backtracking that separates isolated repair from local bursts of moderate-to-severe revision; (ii) evidence on 6{,}000 Qwen3-8B AIME traces that incorrect traces remain more revision-heavy and initiate more late bursts after relative-progress alignment, with the same qualitative asymmetry across 11 additional model--dataset pairs; and (iii) a prefix-causal selective early-exit policy that improves over fixed shallow budgets at 2k--8k words while using only prefix-available features.

\section{Related Work}
\label{sec:related_work}

Early exit and adaptive computation have been studied in several neighboring settings. In deep networks and encoder-style NLP models, classic work showed that easy inputs can often be handled with less computation by using intermediate predictions or confidence-based exits \cite{lee2015deeplysupervised,graves2016adaptive,teerapittayanon2016branchynet,xin2020deebert,liu2020fastbert,zhou2020bertmlosespatience}. For decoder-only LLMs, later work emphasized that practical gains depend not only on prediction quality but also on systems constraints such as batched autoregressive decoding, KV-cache reuse, and verification cost \cite{delcorro2023skipdecode,bae2023free,chen2023eellm,pan2024eetuning,elhoushi2024layerskip,xia2024swift,liu2024kangaroo,fan2024notalllayers,vincenti2024dynamicvocab,kumar2025helios}. Most of this literature, however, focuses on reducing \emph{computation per token} by adapting depth or drafting strategy.

A partially orthogonal line asks when a long reasoning trace itself can be stopped. Recent work has considered answer stabilization, confidence at reasoning transition points, verifier-based stopping, hidden-state probes, and uncertainty-based criteria as signals that further reasoning is unlikely to help \cite{yang2025deer,liu2025answerconvergence,wei2025stopspinningwheels,jiang2025flashthink,chen2025seal,zhang2025reasoningmodelsknow,sharma2025thinkjustenough}. Related approaches make reasoning length more directly controllable through budget conditioning or post-training objectives that reward shorter successful traces \cite{han2024tokenbudgetaware,dai2025sgrpo,aggarwal2025l1}. These papers collectively show that long reasoning often contains redundancy, but they leave open a central descriptive question: which \emph{trace-level dynamics} distinguish useful continued search from unproductive overthinking?

Our work instead uses the internal shape of revision as the signal. We do not add a draft model, verifier, or training objective; we characterize timing, severity, and \emph{burst structure}, then instantiate them in a prefix-causal early-exit policy. This complements prior early-stopping work with a structured trace-level signal for reasoning instability.

\section{Data, Annotation, and Evaluation Setup}
\label{sec:setup}

\subsection{Trace Corpus and Protocol}
Our study uses 6{,}000 reasoning traces from 60 AIME problems spanning AIME2024 and AIME2025 \cite{huggingfaceh4aime2024,opencompass2025aime}, with 30 problems per year and 100 sampled traces per problem. All traces were generated with Qwen3-8B \cite{yang2025qwen3} at temperature 0.6 with explicit thinking enabled. Table~\ref{tab:corpus-summary} summarizes the corpus. Baseline accuracy below is exact-match accuracy on the raw traces; because every question contributes exactly 100 traces, these dataset-level values are also the mean per-question accuracies.

\begin{table}[t]
\centering
\scriptsize
\setlength{\tabcolsep}{3pt}
\renewcommand{\arraystretch}{1.0}
\resizebox{\columnwidth}{!}{%
\begin{tabular}{@{}lrrrrrrr@{}}
\toprule
\textbf{Split} & \textbf{\#Q} & \textbf{\#Traces} & \textbf{Tr/Q} &
\textbf{Base acc.} & \textbf{Mean len.} & \textbf{Med. len.} &
\textbf{Med. segs/tr.} \\
\midrule
AIME2024 & 30 & 3000 & 100 & 73.2\% & 8,743 & 7,309 & 330 \\
AIME2025 & 30 & 3000 & 100 & 64.5\% & 9,646 & 9,288 & 405 \\
\bottomrule
\end{tabular}%
}
\caption{Corpus summary. Baseline accuracy is exact-match accuracy on all raw traces. Length statistics are in words. Segment statistics are computed on the macrostep-annotated subset.}
\label{tab:corpus-summary}
\end{table}

Unless otherwise noted, all length and depth statistics in this work are measured in \emph{words} rather than tokens. This choice follows the macrostep-annotated traces, which store consistent per-segment word counts over the reasoning portion of each output. Reasoning-length and segment-count statistics are computed on the macrostep-annotated traces, while baseline accuracy is computed on the raw traces; the corpus contains 3{,}000 traces per year, or 6{,}000 in total.

The preprocessing protocol is fixed across the corpus. Trace generation uses thinking-enabled sampling, while the later macrostep labeling pass operates with thinking disabled and segments only the reasoning portion of the saved trace. For each corpus, the labeler is the same base model used for trace generation, run in non-thinking mode (e.g., Qwen3-8B for the primary AIME corpus). Segmentation is line based: each non-empty line in the reasoning trace becomes one segment, and each segment is stored with start and end character offsets and a word count. Final-answer extraction uses the last boxed answer when present and otherwise falls back to a case-insensitive numeric match on ``final answer:'' or ``answer:''. Predicted and gold answers are lightly normalized before exact comparison by removing commas, dollar signs, text-wrapper commands, and trailing periods. All accuracy numbers in the work use this final-answer extraction rather than any intermediate state.

\subsection{Segmentation and Backtrack Annotation}
Each segment is assigned one move label from four categories: \emph{continue}, \emph{stall}, \emph{backtrack}, and \emph{exit}, together with a confidence distribution over those moves. The present work uses the backtrack confidence as a severity score; Appendix~\ref{app:severity_scale} anchors the semantic interpretation of this scale. A segment may therefore have the move label \emph{backtrack} without qualifying as an event at a given threshold, and the threshold itself is applied to the backtrack-confidence score rather than to the argmax move label. Throughout the work, a \emph{qualifying backtrack event} means a scored segment whose backtrack-confidence score meets the threshold under consideration.

Burst structure is defined over these qualifying events. Let $d_1 < d_2 < \cdots < d_N$ denote the start depths of qualifying backtrack events measured in reasoning words. Successive qualifying events belong to the same burst when their start-depth gap is at most 500 words. A \emph{multi-burst} contains at least two qualifying events. The whole-trace burst analyses then summarize how often these clusters occur, how large they become, and what fraction of qualifying events fall inside them.

\begin{figure}[t]
\centering
\includegraphics[width=\columnwidth]{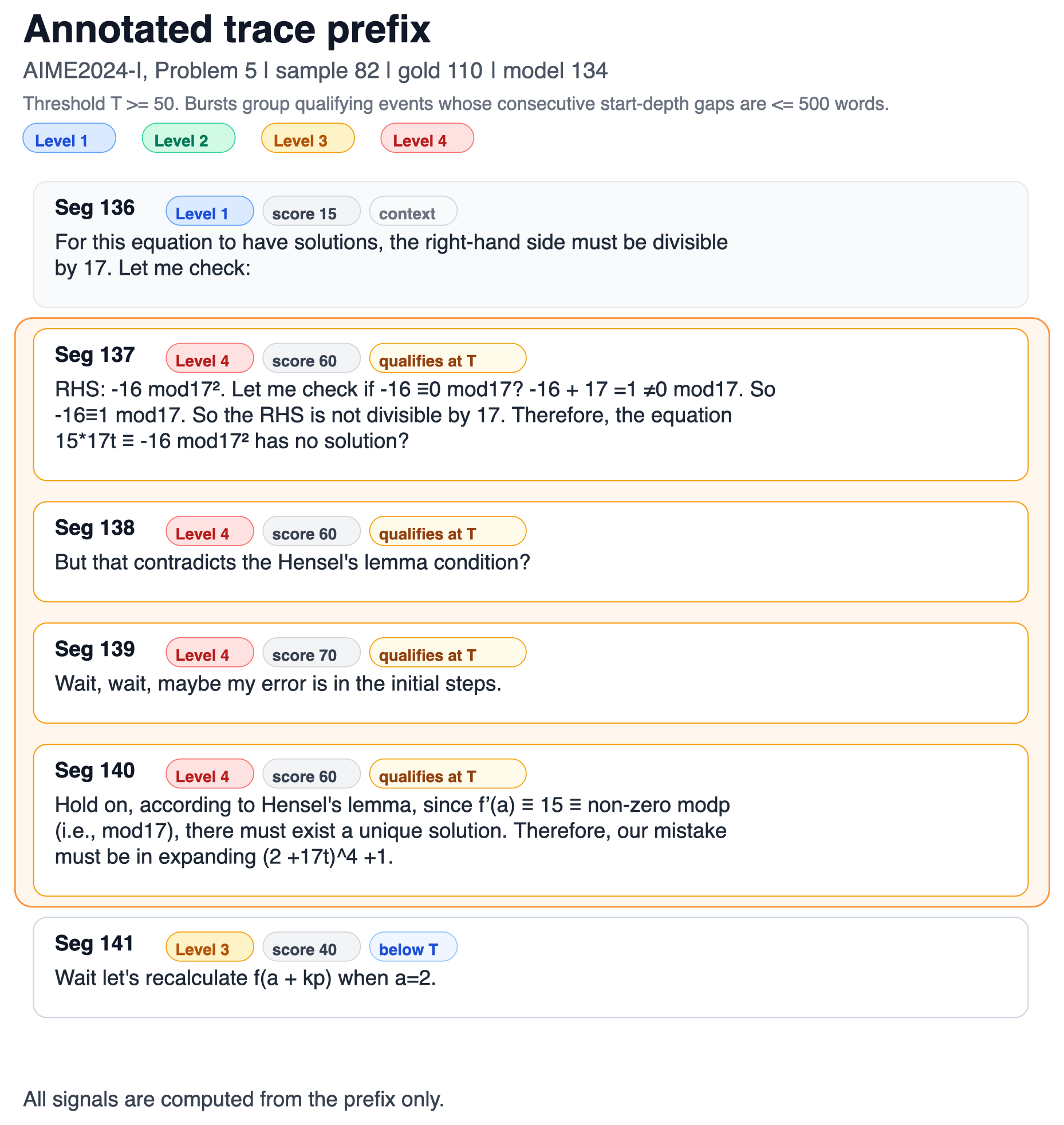}
\caption{Annotated trace prefix from the corpus in threshold view. Segments 137--140 qualify as backtrack events at $T \ge 50$ and form one burst because consecutive qualifying start-depth gaps are at most 500 words. Segment 141 is labeled \emph{backtrack} but is below threshold. The example shows that the useful signal is clustered moderate-to-severe reversal, not merely the presence of any backtrack.}
\label{fig:annotated-trace}
\end{figure}

Figure~\ref{fig:annotated-trace} illustrates the annotation scheme on a real trace prefix. The example highlights three properties that are central to the work's later analyses: segment boundaries are line based, backtrack severity is thresholded rather than binary, and the operationally useful signal is often not a single reversal but a short local cluster of moderate-to-severe reversals.

\subsection{Prefix Availability and Offline Diagnostics}
We separate deployable prefix features from offline diagnostics. Prefix-causal features include current backtrack scores, whether qualifying events have occurred, raw first-event depth, and burst statistics available before the current depth. Diagnostics that normalize by total trace length, such as 20-bin relative-position plots, require completed traces and are not used by the online policy.

\subsection{Evaluation Metrics}
We report exact-match accuracy on raw traces and, for filters, retained-trace accuracy with coverage or drop rate. Timing analyses use event rate and median first qualifying depth. Relative-position analyses split each completed trace into 20 equal-width progress bins and compute qualifying-event probability per eligible segment; sparse-bin support details are deferred to Appendix~\ref{app:filter_setup}. Burst analyses report multi-burst count, maximum burst size, and share of qualifying events inside multi-bursts.

\section{Backtracking Dynamics in Completed Traces}
\label{sec:backtracking_dynamics}

This section moves from a qualitative view of backtrack severity to simple length baselines, then to first-event timing and matched-progress comparisons, and finally to burst structure. Full per-question breakdowns, grouped count views, and auxiliary rejection-style analyses are deferred to the appendix.

\subsection{Qualitative Spectrum, Threshold Regimes, and Burst Definitions}

Backtracking is heterogeneous. At the low end, the model often signals that a local calculation should be checked again, but the correction remains limited in scope. Mid-severity backtracks more often acknowledge a real derivational problem or a need to redo part of the reasoning. High-severity backtracks are more decisive and are frequently tied to contradiction, abandonment of the current line of attack, or explicit re-derivation after apparent progress. This distinction matters because a reasoning model that never revisits an intermediate claim is not necessarily more efficient in any useful sense; some early repair is part of healthy search.

For the quantitative analyses below, the threshold sweep is grouped into four work-level regimes: Level~1 uses $T\ge10$, Level~2 uses $T\ge20,30,40$, Level~3 uses $T\ge50,60$, and Level~4 uses $T\ge70$. A \emph{dense burst} extends the multi-burst notion of \S\ref{sec:setup} by requiring at least three qualifying events. Additional annotated examples, exact threshold-level breakdowns, and auxiliary qualitative material are reported in the appendix.

\paragraph{Burst statistics.}
Fix an exact threshold $\tau$ and one completed trace. Let
\[
d_1 < d_2 < \cdots < d_n
\]
be the start depths, measured in thinking words, of the qualifying backtrack events in that trace, where a qualifying event is a scored segment whose backtrack-confidence score is at least $\tau$. Let $g=500$ denote the burst-gap parameter; a sensitivity check on this choice is reported in Appendix~\ref{app:burst_gap_robustness}. Two consecutive qualifying events belong to the same burst whenever
\[
d_i - d_{i-1} \le g.
\]
This partitions the $n$ qualifying events into $J$ bursts
\[
B_1, B_2, \dots, B_J,
\]
where the size of burst $B_j$ is
\[
m_j \deq |B_j|.
\]
A \emph{multi-burst} is a burst with $m_j \ge 2$, and a \emph{dense burst} is a burst with $m_j \ge 3$.

From this partition we define four per-trace burst statistics. We write $K_{\ge 2}$ for the number of multi-bursts, $S_{\ge 2}$ for the share of qualifying events that lie inside multi-bursts, $\rho$ for the compression ratio, and $m_{\max}$ for the maximum burst size:
\[
K_{\ge 2}
\deq
\sum_{j=1}^{J} \mathbf{1}[m_j \ge 2],
\]
\[
S_{\ge 2}
\deq
\frac{\sum_{j=1}^{J} m_j \mathbf{1}[m_j \ge 2]}{n},
\]
\[
\rho
\deq
\frac{n}{J},
\]
\[
m_{\max}
\deq
\max_{1 \le j \le J} m_j.
\]
When a trace has no qualifying events, these burst statistics are set to zero.

Dense-burst analogues use the same notation with $\ge 2$ replaced by $\ge 3$, for example
\[
K_{\ge 3}
\deq
\sum_{j=1}^{J} \mathbf{1}[m_j \ge 3].
\]

For each class $c \in \{\textnormal{correct}, \textnormal{wrong}\}$, the burst figures report class means of these per-trace quantities:
\[
\bar f^{(c)}
\deq
\frac{1}{|\mathcal{T}_c|}\sum_{T \in \mathcal{T}_c} f(T),
\qquad
f \in \{K_{\ge 2}, S_{\ge 2}, \rho, m_{\max}\},
\]
where $\mathcal{T}_c$ is the set of traces in class $c$.

\subsection{Simple Baselines}

Simple baselines remain strong, and it is useful to establish that before turning to richer temporal and structural analyses. Table~\ref{tab:backtracking-word-baselines} summarizes a compact reasoning-length sweep. The implemented cutoff uses total output word count, so the table should be read as a coarse reasoning-length baseline rather than as a tokenizer-level latency result.

Even this simple baseline moves retained accuracy substantially. With no limit, baseline accuracy is 73.2\% on AIME2024 and 64.5\% on AIME2025. A 12{,}000-word limit raises retained accuracy to 80.6\% and 71.7\%, with drop rates of 29.9\% and 38.7\%, respectively. A 10{,}000-word limit still reaches 82.9\% on AIME2024 and 68.0\% on AIME2025, but at more aggressive drop rates of 38.6\% and 50.0\%. Once the limit becomes too aggressive, however, the retained pool deteriorates quickly: at 8{,}000 words, retained accuracy falls to 71.9\% on AIME2024 and 54.4\% on AIME2025. The implication is not that length alone solves the problem, but that wrong traces do tend to accumulate more burden and more extended reasoning overall. Any stronger structural signal therefore has to outperform a fairly competitive coarse baseline.

\begin{table}[t]
\centering
\footnotesize
\setlength{\tabcolsep}{2.5pt}
\renewcommand{\arraystretch}{1.0}
\resizebox{\columnwidth}{!}{%
\begin{tabular}{@{}rcccc@{}}
\toprule
{\scriptsize\textbf{Word limit}} &
{\scriptsize\textbf{AIME2024 accuracy}} &
{\scriptsize\textbf{AIME2025 accuracy}} &
{\scriptsize\textbf{AIME2024 drop rate}} &
{\scriptsize\textbf{AIME2025 drop rate}} \\
\midrule
No limit & 73.2\% & 64.5\% & 0.0\% & 0.0\% \\
16{,}000 & 76.4\% & 66.2\% & 13.0\% & 13.4\% \\
12{,}000 & 80.6\% & 71.7\% & 29.9\% & 38.7\% \\
10{,}000 & 82.9\% & 68.0\% & 38.6\% & 50.0\% \\
8{,}000  & 71.9\% & 54.4\% & 49.9\% & 61.5\% \\
\bottomrule
\end{tabular}%
}
\caption{Word-limit baselines for retained-trace accuracy. Moderate limits improve retained accuracy before year-specific drop rates become too aggressive.}
\label{tab:backtracking-word-baselines}
\end{table}

\subsection{Timing and Relative Position}

\begin{figure}[t]
\centering
\includegraphics[width=\columnwidth]{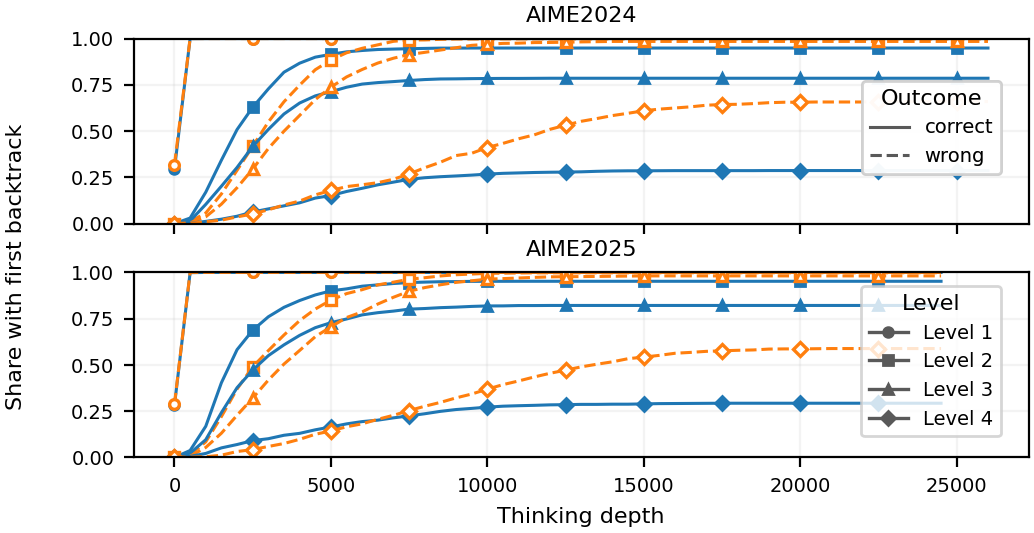}

\caption{First-backtrack timing curves by year, grouped into the four work-level regimes. The work-level pattern is that the middle levels reach their first qualifying event earlier in correct traces, while wrong traces usually end at higher event coverage.}
\label{fig:first-backtrack-timing-panels}
\end{figure}

Timing and relative position matter because they separate benign early revision from the later instability that looks more like overthinking. Figure~\ref{fig:first-backtrack-timing-panels} shows the first-backtrack timing pattern. The loosest regime is nearly universal and therefore weakly discriminative, while the extreme regime is informative but sparse. The most useful separation appears in the middle regimes. There, the median first qualifying backtrack usually appears \emph{earlier} in correct traces than in wrong traces, even though wrong traces cover more of the corpus at those same levels.

At Level~2, the median first qualifying backtrack in AIME2024 occurs at $\sim$ 2000 words for correct traces but at $\sim$ 3000 words for wrong traces. In AIME2025, the corresponding medians are $\sim$ 1700 versus $\sim$ 2500 words. The same pattern persists at Level~3: in AIME2024, correct traces reach their first qualifying event at $\sim$ 2500 words compared with $\sim$ 3500 words for wrong traces, while in AIME2025 the corresponding values are $\sim$ 2100 versus $\sim$ 3500 words. Yet wrong traces still cover more of the corpus at those same levels. At Level~2, event rates are 93.0--95.8\% for correct traces versus 99.8--99.9\% for wrong traces in AIME2024, and 94.1--95.8\% versus 99.8\% in AIME2025. At Level~3, the event rates are 77.1--80.0\% versus 98.1--98.8\% in AIME2024 and 80.8--83.5\% versus 97.7--98.5\% in AIME2025. The timing curves therefore argue against the naive rule that any early backtrack signals failure. Early local repair is often compatible with success; what distinguishes wrong traces is broader and later persistence.

\begin{figure}[t]
\centering
\includegraphics[width=\columnwidth]{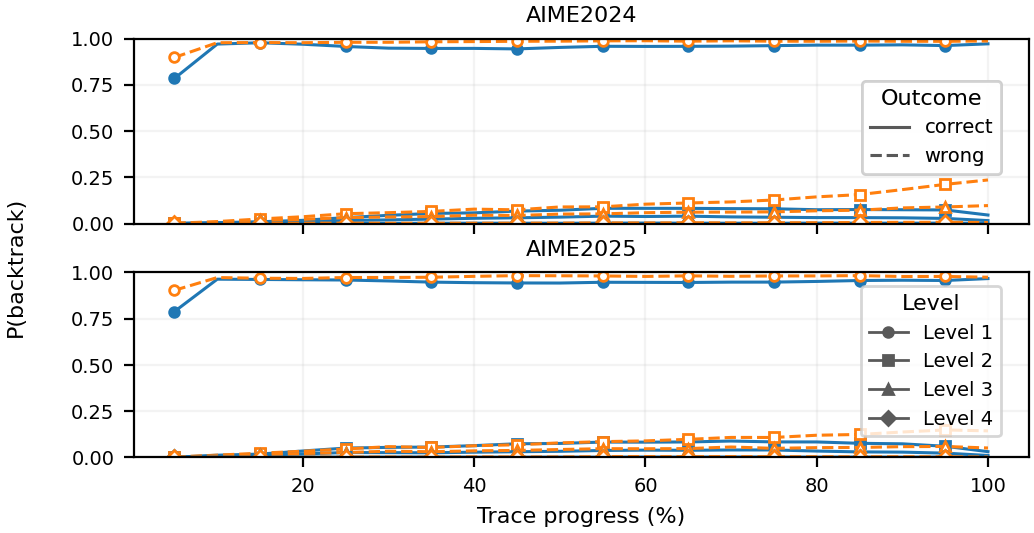}

\caption{Normalized-depth probability plots by year. After matching traces by relative progress rather than raw word depth, wrong traces still occupy more revision-heavy states across most bins.}
\label{fig:normalized-depth-probability-panels}
\end{figure}

Figure~\ref{fig:normalized-depth-probability-panels} then removes the raw-length explanation by aligning completed traces by relative progress rather than absolute word depth. After this alignment, wrong traces remain more backtrack-heavy across most bins, especially in the middle regimes and especially late in the trajectory. At Level~2, the last-bin normalized backtrack probability in AIME2024 is 0.040--0.050 for correct traces but 0.213--0.247 for wrong traces; in AIME2025, the corresponding values are 0.027--0.034 versus 0.130--0.151. Level~3 shows the same qualitative pattern at lower absolute rates: the last-bin probability in AIME2024 is 0.016--0.017 for correct traces and 0.095--0.099 for wrong traces, while in AIME2025 it is 0.010--0.012 versus 0.050--0.053. The normalized view therefore shows that the separation is not only a side effect of wrong traces being longer. Even at matched progress, wrong traces spend more of the late trajectory in revision-heavy states. The grouped count views and the normalized trace-rejection extension are kept in the appendix because they are supplementary to the central timing result.


\subsection{Burst Structure}

\begin{figure*}[t]
\centering
\includegraphics[width=0.98\textwidth]{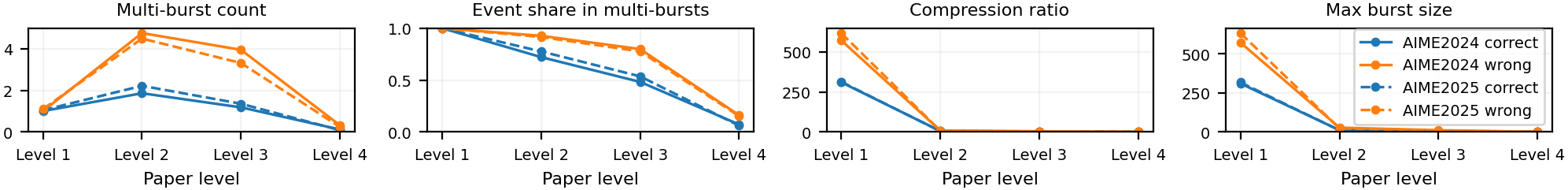}
\caption{Grouped whole-trace burst metrics by year. The important pattern is that the middle levels consistently separate correct and wrong traces on multi-burst count, event share in multi-bursts, and maximum burst size. The compression subplot is shown for completeness, but it is less interpretable in the loosest regime because long chains dominate there.}
\label{fig:burst-structure-panels}
\end{figure*}

Timing already suggests that wrong traces differ less by the mere appearance of revision than by its later persistence. Burst analyses sharpen that observation by asking whether later revisions remain isolated or instead collapse into repeated nearby clusters. Figure~\ref{fig:burst-structure-panels} shows that wrong traces do not simply accumulate more backtrack events in aggregate; they also organize those events into more repeated local bursts. The loosest regime is too permissive to be useful, because almost every qualifying event is absorbed into a long chain. From Level~2 onward, however, the separation is strong in both years.

At Level~2, the average multi-burst count in AIME2024 rises from 1.793--1.897 in correct traces to 4.711--4.776 in wrong traces, while the share of events occurring inside multi-bursts rises from 0.695--0.734 to 0.922--0.929. In AIME2025, the corresponding ranges are 2.140--2.260 versus 4.427--4.522 for multi-burst count, and 0.756--0.787 versus 0.909--0.919 for the share of events inside multi-bursts. Level~3 shows the same qualitative pattern at lower absolute rates. In AIME2024, multi-burst count rises from 1.154--1.215 in correct traces to 3.911--3.989 in wrong traces, while event share in multi-bursts rises from 0.470--0.496 to 0.794--0.804. In AIME2025, the corresponding values are 1.293--1.424 versus 3.266--3.363 and 0.515--0.557 versus 0.771--0.781. The whole-trace view therefore strengthens the work's central claim: a single reversal can still be healthy, but repeated local clustering is much harder to read as productive search.


\begin{figure*}[t]
\centering
\includegraphics[width=0.98\textwidth]{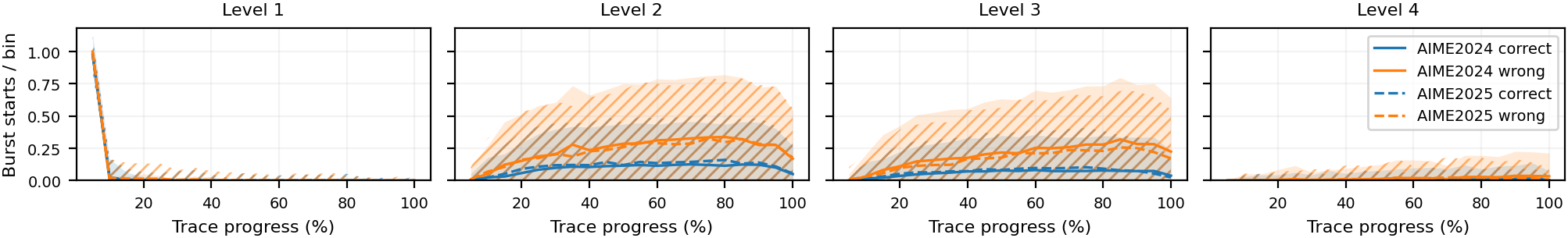}
\caption{Grouped normalized burst-start count plots by year. After aligning traces by relative progress, wrong traces still initiate more multi-burst clusters across most bins, especially in the middle levels.}
\label{fig:burst-normalized-panels}
\end{figure*}

Figure~\ref{fig:burst-normalized-panels} shows that this pattern survives matched-progress comparison. After aligning completed traces by relative progress, wrong traces still initiate more multi-burst clusters later in the trajectory. At Level~2, the last-bin normalized burst-start count in AIME2024 rises from 0.046--0.052 for correct traces to 0.160--0.174 for wrong traces; in AIME2025, the corresponding values are 0.045--0.059 versus 0.174--0.180. Level~3 again shows the same qualitative pattern at lower absolute rates: the last-bin count in AIME2024 is 0.038--0.039 for correct traces and 0.222--0.227 for wrong traces, while in AIME2025 it is 0.024--0.026 versus 0.170--0.174. The normalized burst-start view is therefore the clearest matched-progress counterpart to the timing result: wrong traces do not just revise later; they are also more likely to restart clustered instability later.

Taken together, these results make burst structure the strongest structural signal in the present study. The middle regimes carry the most stable separation: the loosest regime captures ubiquitous checking behavior, while the sparsest regime is informative but too rare to serve as the main descriptive picture. Auxiliary single-signal and likelihood-ratio analyses for burst-based rejection are reported in Appendix~\ref{app:additional_evidence_structural_signal}, and a sensitivity check on the burst-gap parameter is reported in Appendix~\ref{app:burst_gap_robustness}.


Across these analyses, simple length filtering is already a reasonable baseline, but timing shows that early correction is not itself a failure signal and the normalized-depth comparison shows the separation is not reducible to length alone. Broader threshold sweeps and auxiliary breakdowns are reported in the appendix.

\section{Burst-Aware Filtering and Prefix-Causal Early Exit}
\label{sec:from_description_to_decision}

Backtracking bursts are not only a descriptive marker of overthinking; they define a deployable prefix-causal policy for selective early exit, with the largest gains over fixed stopping at shallow and intermediate reasoning depths.

Having established that late backtracking bursts mark unstable reasoning, we next turn the signal into a prefix-causal filtering policy. The policy is deployable in the inference-time sense: at a fixed checkpoint $d$, it computes only features available before $d$, scores the trace for likely failure, and exits or triages traces whose score exceeds a learned cutoff. We evaluate the policy by replaying this decision rule over completed traces, which lets us compare retained accuracy and drop rate against matched fixed-depth baselines. Thus the section studies a deployable rule under retrospective evaluation, not merely an offline diagnostic.

\subsection{Completed-Trace Filtering}
\label{sec:completed_trace_filtering}

Table~\ref{tab:completed_filtering_results} compares representative completed-trace rules. The headline result is that burst-aware structure wins at aggressive operating points: at $\sim$50\% drop, the burst-only filter (47--56\% drop) reaches 77.8\% / 68.0\% retained accuracy, while the matched hard 8k cutoff (50--62\% drop) reaches only 71.9\% / 54.4\% --- a $+5.9$ / $+13.6$ point gap on AIME2024/2025. The structural filters also avoid the catastrophic failure of harsher truncation: a 5k cutoff collapses retained accuracy to 45.7\% / 35.9\% (Appendix~\ref{app:hard_cutoff_sweep}). At more conservative drop rates the picture flips: a 12k limit reaches 80.6\% / 71.7\% and a 10k limit reaches 82.9\% / 68.0\%, neither of which is surpassed by the learned structural filters (completed hybrid 77.4\% / 65.0\%; burst-only 77.8\% / 68.0\%; formal definitions in Appendix~\ref{app:completed_hybrid} and Appendix~\ref{app:burst_only_fallback}).

This comparison sharpens the claim of the paper. The structural filters are not the best completed-trace decision rule once a fairly late hard cutoff is allowed, but they dominate harsher truncation at matched aggressiveness and they separate useful structure from blunt truncation. The hybrid is therefore best interpreted as the most faithful operationalization of the paper's descriptive story, not as the strongest practical full-trace filter. The structural filters win over harsh fixed cutoffs at aggressive operating points, but moderate hard cutoffs and some burden-only ablations remain competitive (Appendix~\ref{app:completed_aux_results}). Thus the completed-trace result should be read as evidence that burst structure carries useful signal, not as evidence that the composite structural filter dominates every simpler rule.

\begin{table}[t]
\centering
\footnotesize
\setlength{\tabcolsep}{3.5pt}
\renewcommand{\arraystretch}{1.02}
\resizebox{\columnwidth}{!}{%
\begin{tabular}{@{}lcccc@{}}
\toprule
\textbf{Method} & \textbf{AIME2024 acc.} & \textbf{AIME2025 acc.} & \textbf{AIME2024 drop} & \textbf{AIME2025 drop} \\
\midrule
No filter & 73.2\% & 64.5\% & 0.0\% & 0.0\% \\
Hard cutoff 12k & 80.6\% & 71.7\% & 29.9\% & 38.7\% \\
Hard cutoff 10k & 82.9\% & 68.0\% & 38.6\% & 50.0\% \\
Hard cutoff 8k & 71.9\% & 54.4\% & 49.9\% & 61.5\% \\
Burst-only & 77.8\% & 68.0\% & 47.4\% & 55.6\% \\
Completed hybrid & 77.4\% & 65.0\% & 44.8\% & 73.0\% \\
\bottomrule
\end{tabular}%
}
\caption{Representative completed-trace filtering results. Moderate hard cutoffs (12k, 10k) remain the strongest full-trace baselines at conservative drop rates. At more aggressive drop rates ($\sim$50\%), the burst-only filter beats the matched hard 8k cutoff by $+5.9$ on AIME2024 and $+13.6$ on AIME2025; burst-aware rules are also much less destructive than harsh truncation.}
\label{tab:completed_filtering_results}
\end{table}

\subsection{Prefix-Causal Early-Exit Policy}
\label{sec:prefix_causal_early_exit}

The prefix-causal setting is the intended deployment point of the method. Unlike completed-trace filters, which observe the final trajectory length or full normalized profile, the prefix policy acts at a fixed checkpoint using only the observed reasoning prefix. This makes it suitable for selective early exit: the system can stop, abstain, resample, or route a trace once the observed backtracking pattern indicates likely failure.

The prefix policy produces large gains precisely where early-exit decisions matter most. At 2k words, a fixed length-retention cutoff is highly destructive at shallow depths, reaching only 10.0\% / 6.7\%, while the prefix policy retains 73.5\% / 64.5\% accuracy and exits 21.7\% / 27.0\% of traces. At 5k, the same pattern holds: fixed truncation reaches 45.7\% / 35.9\%, whereas the prefix policy reaches 68.2\% / 60.2\%. The advantage remains positive at 8k and narrows only at 12k, where a late fixed cap becomes a strong baseline. Thus the method's deployment niche is selective early exit at shallow and intermediate depths, not replacement of every possible late length cap.

\begin{algorithm}[t]
\caption{Prefix-Causal Burst Early Exit}
\label{alg:prefix_early_exit}
\textbf{Input:} observed reasoning prefix up to depth $d$; trained prefix score $f_d$; cutoff $c_d$; qualifying-event thresholds.
\begin{enumerate}
\item Segment the observed prefix.
\item Compute prefix-available burden, burst, and timing features: $r_{20}(d)$, $S_{\ge 2}^{(20)}(d)$, $\rho^{(40)}(d)$, $m_{\max}^{(20)}(d)$, $h_{50}(d)$, $b_{40}(d)$.
\item Compute $p_{\mathrm{wrong}}^{\mathrm{prefix}}(d) = f_d(\mathbf{x}(d))$.
\item At $d = 2000$, continue if $N_{20}(d) < 2$.
\item If $p_{\mathrm{wrong}}^{\mathrm{prefix}}(d) \ge c_d$, flag the trace for early termination or triage.
\item Otherwise, continue generation to the next checkpoint.
\end{enumerate}
\end{algorithm}

In the experiments below, a flagged trace is treated as exited from the retained set; retained accuracy is computed over unflagged traces.

All features ($r_{20}(d)$, $S_{\ge 2}^{(20)}(d)$, $\rho^{(40)}(d)$, $m_{\max}^{(20)}(d)$, $h_{50}(d)$, $b_{40}(d)$) are functions of the observed prefix only --- no completed-trace normalization or future backtrack event is used --- and are defined formally in Appendix~\ref{app:online_prefix_filter}; matched-depth ablations against burden-only prefix variants are reported in Appendix~\ref{app:prefix_aux_results}. The 2k guard ($N_{20}(2000) \ge 2$, step 4) encodes the timing finding from Section~\ref{sec:backtracking_dynamics} that a single early correction is not itself a failure signature.

These results identify the practical role of burst-aware control. A fixed shallow length filter treats all long traces alike and therefore removes many traces that would have recovered through early repair. The prefix policy instead exits selectively, using accumulated burden, timing, and burst structure to distinguish recoverable repair from likely instability. Its strongest operating region is 2k--8k words, where it substantially improves over fixed length-based filtering while still using only prefix-available information.

\begin{table}[t]
\centering
\footnotesize
\setlength{\tabcolsep}{3pt}
\renewcommand{\arraystretch}{1.02}
\resizebox{\columnwidth}{!}{%
\begin{tabular}{@{}lccc@{}}
\toprule
\textbf{Prefix depth} & \textbf{Hard cutoff acc.} & \textbf{Online prefix acc.} & \textbf{Online prefix drop} \\
 & \textbf{(A24 / A25)} & \textbf{(A24 / A25)} & \textbf{(A24 / A25)} \\
\midrule
2k & 10.0 / 6.7 & 73.5 / 64.5 & 21.7 / 27.0 \\
5k & 45.7 / 35.9 & 68.2 / 60.2 & 64.6 / 49.1 \\
8k & 71.9 / 54.4 & 73.9 / 64.1 & 42.7 / 45.2 \\
12k & 80.6 / 71.7 & 79.4 / 69.3 & 26.8 / 35.0 \\
\bottomrule
\end{tabular}%
}
\caption{Prefix-causal filtering versus matched hard cutoffs. Trace-aware early exit is most useful at shallow and intermediate depths, where fixed truncation is especially blunt.}
\label{tab:prefix_filtering_results}
\end{table}

\section{Cross-Corpus Generality}
\label{sec:cross_corpus_replication}

The preceding section uses Qwen3-8B on AIME as the primary corpus because it provides repeated samples per problem and long reasoning traces. To test whether the descriptive pattern is specific to that setting, we repeat the same single-corpus analysis on additional model and domain pairs. This comparison is deliberately not a leaderboard: the corpora differ in sampling budget (AIME provides 100 traces per question, GPQA \cite{rein2023gpqa} provides 10, while OmniMath \cite{gao2024omnimath} provides one trace per question, which makes per-question statistics noisier and renders the leave-one-question prefix protocol degenerate), answer format, difficulty distribution, and correctness labeling. Its purpose is narrower. We ask whether the qualitative signature of the primary corpus---wrong traces being longer and more locally bursty at middle backtrack thresholds---continues to appear under model and domain shifts.

\begin{table*}[t]
\centering
\scriptsize
\setlength{\tabcolsep}{3.5pt}
\renewcommand{\arraystretch}{1.03}
\resizebox{0.98\textwidth}{!}{%
\begin{tabular}{@{}llrrrrrr@{}}
\toprule
\textbf{Corpus} & \textbf{Model} & \textbf{\#Tr.} & \textbf{\#Q/Items} &
\textbf{Base acc.} & \textbf{Med. len. C/W} &
\textbf{$T\!\ge\!20$ multi-bursts C/W} &
\textbf{$T\!\ge\!50$ multi-bursts C/W} \\
\midrule
AIME2024 & Qwen3-8B & 3000 & 30 & 73.2\% & 5{,}752 / 14{,}898 & 1.90 / 4.78 & 1.21 / 3.99 \\
AIME2024 & Phi4R & 300 & 30 & 72.7\% & 1{,}823 / 11{,}555 & 0.62 / 2.35 & 0.56 / 2.06 \\
AIME2024 & Qwen3.5-9B & 300 & 30 & 93.3\% & 9{,}432 / 26{,}563 & 2.61 / 4.60 & 4.67 / 10.10 \\
\midrule
AIME2025 & Qwen3-8B & 3000 & 30 & 64.5\% & 6{,}385 / 14{,}627 & 2.26 / 4.52 & 1.42 / 3.36 \\
AIME2025 & Phi4R & 300 & 30 & 68.0\% & 2{,}560 / 10{,}712 & 0.58 / 2.45 & 0.48 / 2.21 \\
AIME2025 & Qwen3.5-9B & 300 & 30 & 83.3\% & 8{,}576 / 26{,}202 & 2.40 / 4.96 & 3.90 / 8.40 \\
\midrule
GPQA-Diamond & Qwen3-8B & 1980 & 198 & 60.6\% & 2{,}220 / 3{,}468 & 0.77 / 1.05 & 0.66 / 0.89 \\
GPQA-Diamond & Phi4R & 1980 & 198 & 66.1\% & 1{,}475 / 5{,}622 & 0.47 / 1.37 & 0.42 / 1.29 \\
GPQA-Diamond & Qwen3.5-9B & 1980 & 198 & 82.2\% & 3{,}296 / 5{,}249 & 1.63 / 2.32 & 1.43 / 2.35 \\
\midrule
OmniMath & Qwen3-8B-thinking & 4428 & 4428 & 66.0\% & 4{,}408 / 11{,}004 & 1.61 / 3.60 & 1.31 / 3.46 \\
OmniMath & Phi4R & 4428 & 4428 & 65.3\% & 1{,}467 / 6{,}010 & 0.52 / 1.64 & 0.46 / 1.43 \\
OmniMath & Qwen3.5-9B & 4428 & 4428 & 70.5\% & 7{,}302 / 16{,}273 & 3.00 / 5.29 & 3.00 / 6.42 \\
\bottomrule
\end{tabular}%
}
\caption{Cross-corpus replication of the length and burst asymmetries. Lengths are median annotated words for correct/wrong traces. Burst columns report class means of the number of multi-bursts per trace. Horizontal rules separate datasets, not model families. OmniMath rows use GPT-4o-judged correctness labels \cite{openai2024gpt4o}, while AIME and GPQA rows use answer extraction or multiple-choice correctness. Threshold regimes (Level~2 = $T\!\ge\!20,30,40$; Level~3 = $T\!\ge\!50,60$) are calibrated to the Qwen3-8B AIME score distribution; the Qwen3.5-9B \cite{qwen2026qwen35modelcard} and Phi4R \cite{abdin2025phi4reasoning} annotators produce markedly different score histograms (e.g., Qwen3.5-9B mass concentrates in $[0,10)$ and $[80,90)$ with a near-empty mid-range), so the cross-corpus claim should be read as ``wrong $>$ correct in burst structure at comparable middle/high severity cuts,'' not as a calibrated threshold comparison.}
\label{tab:cross_corpus_signature}
\end{table*}

Table~\ref{tab:cross_corpus_signature} shows that the central signature is not confined to the primary Qwen3-8B AIME setting. Across the additional long-form reasoning corpora, wrong traces have larger median annotated length than correct traces and larger mean multi-burst counts in both the middle and high-middle regimes. The magnitude is not uniform. The separation is strongest on AIME and OmniMath, where long mathematical traces leave more room for repeated local instability. It is weaker on GPQA-Diamond, whose traces are shorter and whose answer format changes the opportunity structure for extended re-derivation, but the direction of the burst asymmetry remains the same in all three GPQA model rows. The Qwen3.5-9B rows in particular illustrate the threshold-calibration caveat in the table caption: on AIME2024 the multi-burst count rises \emph{non-monotonically} with threshold (1.64/2.00 at $T\!\ge\!10$, 2.61/4.60 at $T\!\ge\!20$, 4.66/10.10 at $T\!\ge\!30$, then flat through $T\!\ge\!70$), a direct reflection of the bimodal score histogram concentrating mass in $[0,30)$ and $[80,100)$ with a nearly empty mid-range, and a reminder that the same numeric threshold does not cut severity uniformly across labelers.

The corresponding Table~\ref{tab:prefix_filtering_results}-style operational grid is reported in Appendix~\ref{app:cross_corpus_prefix_details}. Its main lesson is not that one prefix rule dominates everywhere. Rather, trace-aware prefix filtering is most useful when a fixed cutoff is too shallow for the corpus: the gap is large for Qwen3.5-9B on AIME and for the judged OmniMath rows, while it is modest on shorter GPQA traces and concentrated at shallow depths for Phi4R AIME. On the judged OmniMath rows the gap is driven entirely by the hard-cutoff side: with one trace per question, the leave-one-question prefix filter has no within-question signal to learn from and retains every trace at every depth, so the online column equals the no-filter baseline by construction rather than by an active stopping decision. On Phi4R AIME the prefix gain is concentrated at the shallow 2k depth (AIME2024 43.9 vs.\ 72.1; AIME2025 26.2 vs.\ 67.8), modest at 5k--8k, and slightly negative at 12k on both years (AIME2024 80.1 vs.\ 78.3; AIME2025 75.2 vs.\ 73.0). The Phi4R AIME prefix benefit is therefore a shallow-depth phenomenon rather than a uniform improvement. This reinforces the narrower interpretation of the decision results: burst-aware structure helps avoid the worst failures of blunt truncation, but the value of a particular stopping policy depends on model, domain, and sampling regime.

The replication results are therefore best read as support for the paper's descriptive vocabulary rather than as a claim that a single stopping rule transfers unchanged. The stable part is the contrast between isolated repair and clustered instability. The corpus-dependent part is how much that contrast improves over simple length and burden baselines. This is why the operational filtering study above remains focused on the primary AIME corpus, where the sampling protocol is most controlled, while the auxiliary corpora are used to check whether the underlying trace-level phenomenon is broader than one model-domain pair.

\section{Conclusion}
\label{sec:discussion_and_limitations}

We studied backtracking dynamics in long reasoning traces and showed that the organization of revision matters. Early isolated repair is often compatible with success, while later clustered backtracking is associated with incorrect reasoning. Burst structure provides both a descriptive marker of reasoning instability and a prefix-causal signal for selective early exit. Moderate length cutoffs remain strong completed-trace baselines, but burst-aware control is most useful when shallow fixed filtering would otherwise discard recoverable reasoning. The cross-corpus replication suggests this asymmetry reflects a property of long reasoning rather than a quirk of any single model--domain pair. Treating the organization of revision as a first-class control axis, rather than only its accumulated length, is a natural next step for inference-time policies in reasoning systems.

The study has several limitations. The most controlled corpus is Qwen3-8B on AIME, while the cross-corpus rows vary in model, domain, sample count, answer format, and judging protocol. They therefore support qualitative replication rather than a fully matched benchmark comparison. Backtrack severity is produced by a model-based annotation pass rather than a human gold standard. Some profile-shape features require completed traces and are offline by construction. The prefix policy is executable online because all of its features are prefix-causal, but the present evaluation replays the policy over completed traces rather than integrating it into a live inference stack. Systems-level latency, batching effects, and downstream choices among stopping, abstention, and resampling remain outside the scope of this study.

\bibliography{references}
\bibliographystyle{icml2026/icml2026}

\appendix

\section{Severity-Scale Sanity Check}
\label{app:severity_scale}

Before defining the filtering rules, it is useful to anchor the backtrack severity scale semantically. Representative samples show that the loosest score range overlaps heavily with ordinary continuation and verification behavior, whereas higher-score buckets are much more clearly associated with explicit reconsideration, contradiction, and re-evaluation. This semantic shift is consistent with the work-level grouping used in the main text: Level~1 is loose, Level~2 is the main middle regime, Level~3 is high-middle, and Level~4 is extreme.

\begin{table*}[!t]
\centering
\footnotesize
\setlength{\tabcolsep}{3pt}
\renewcommand{\arraystretch}{1.03}
\resizebox{\textwidth}{!}{%
\begin{tabular}{@{}lrrrrrl@{}}
\toprule
\textbf{Bucket} & \textbf{Segments} & \textbf{Correct} & \textbf{Wrong} & \textbf{Median score} & \textbf{Median words} & \textbf{Common summary terms} \\
\midrule
0--10 & 79,173 & 54,573 & 24,600 & 4.0 & 14.0 & equation, compute, calculate, using, verify, problem \\
40--50 & 84,211 & 37,853 & 46,358 & 40.0 & 23.0 & verify, calculation, evaluate, approach, check, need \\
60--70 & 71,348 & 30,966 & 40,382 & 60.0 & 27.5 & verify, need, approach, evaluate, calculation, check \\
70--80 & 5,816 & 2,588 & 3,228 & 70.0 & 25.0 & verify, approach, previous, need, evaluate, contradiction \\
\bottomrule
\end{tabular}%
}
\caption{Selected representative-sample buckets used to interpret the backtrack severity scale. Higher-score buckets are increasingly dominated by explicit reconsideration rather than ordinary continuation.}
\label{tab:severity_scale_sanity_check}
\end{table*}

\paragraph{Per-segment signal is weak; the lift comes from aggregation.}
The per-segment severity score is only weakly predictive of trace-level correctness. On AIME2024, the wrong-segment rate in the high-severity buckets stays well below 100\% (55.2\% in $[60,70)$ and 51.6\% in $[70,80)$), against a baseline wrong-segment rate of $26.8\%$ on this corpus. So a single high-severity segment is not a reliable failure signal on its own; the descriptive power exploited throughout the main text comes from aggregating these per-segment scores into counts, first-event timing, and bursts, not from the per-segment label.

\section{Filtering Setup and Notation}
\label{app:filter_setup}

All filtering methods start from the same object: a completed or prefix-truncated reasoning trace with macrostep labels, backtrack scores, and word-depth positions.

Let $D$ denote the usable depth of the trace in reasoning words. For completed-trace filters, $D$ is the full trace length. For prefix-causal filters, $D=d$ is a fixed observed prefix depth such as $2000$, $5000$, $8000$, or $12000$.

For a threshold $T$, let $N_T(D)$ be the number of qualifying backtrack events whose backtrack-confidence score is at least $T$ and whose start depth is less than $D$:
\[
N_T(D)
\deq
\sum_i \mathbf{1}\!\left[s_i \ge T \;\wedge\; d_i < D\right].
\]

The simplest burden feature is the thresholded backtrack rate per 1000 words:
\[
r_{20}(D)
\deq
1000 \cdot \frac{N_{20}(D)}{D}.
\]

For full-trace profile features, relative progress is divided into $B=20$ equal-width bins. If $C_b(T)$ is the number of score-$\ge T$ events in bin $b$ and $M_b$ is the number of scored segments in that bin, then the per-bin backtrack probability is
\[
p_b(T) \deq \frac{C_b(T)}{M_b}.
\]
We also write
\[
p(T) \deq \bigl(p_1(T),\dots,p_{20}(T)\bigr)
\]
for the corresponding 20-dimensional profile vector. For the per-bin probabilities reported in Section~\ref{sec:backtracking_dynamics}, we further apply a minimum per-class support filter to stabilize estimates against sparse classes in extreme bins: correct $n=2{,}178$, wrong $n=805$ on AIME2024; correct $n=1{,}923$, wrong $n=1{,}057$ on AIME2025. Recomputing without this filter from the raw per-segment data yields the same qualitative ordering but slightly different absolute rates.

Burst features use the same construction defined in Section~\ref{sec:setup}: qualifying events at threshold $T$ are partitioned into bursts $B_1,\dots,B_J$ with sizes $m_j \deq |B_j|$ using burst-gap $g=500$ words. From this partition we use three per-trace summaries:
\[
S_{\ge 2}
\deq
\frac{\sum_{j=1}^{J} m_j \mathbf{1}[m_j \ge 2]}{n},
\quad
\rho
\deq
\frac{n}{J},
\quad \]
\[
m_{\max}
\deq
\max_j m_j.
\]
Here $S_{\ge 2}$ is the share of qualifying events inside multi-bursts, $\rho$ is the compression ratio, and $m_{\max}$ is the maximum burst size.

All learned filters use the same final scoring template. If $x_j$ are the raw features and
\[
z_j \deq \frac{x_j - \mu_j}{\sigma_j}
\]
denotes training-split standardization, then the wrong-trace score is
\[
p_{\mathrm{wrong}}
\deq
\sigma\!\left(\beta_0 + \sum_j \beta_j z_j\right),
\]
where $\sigma(\cdot)$ is the logistic sigmoid. A trace is kept when
\[
p_{\mathrm{wrong}} < c,
\]
and the cutoff $c$ is chosen on the training questions to maximize mean per-question retained accuracy.

All learned filters are evaluated in a question-controlled way. When scoring traces from question $q$, every trace from $q$ is held out from model fitting, feature standardization, cutoff selection, and any reference-profile construction. This avoids question-level leakage and makes the reported retained-accuracy numbers genuinely out-of-question.

\section{Filtering Methods}
\label{app:filter_methods}

All filters below take a completed or prefix-truncated trace with macrostep labels and produce a keep/drop decision. The hard length cutoff baseline keeps a trace when its reasoning length satisfies $D \le L$ for a fixed limit $L$; no trace content beyond depth is used. The remaining filters introduce structural and burden features.

\subsection{Completed Hybrid}
\label{app:completed_hybrid}

The completed hybrid is the most faithful implementation of the paper's structural story because it combines \emph{where} instability happens with \emph{how} it is organized.

The first three features describe the normalized shape of backtracking over full-trace progress.

\paragraph{Profile slope at $T \ge 20$.}
Let the normalized bin centers be
\[
x_b \in \{0.025, 0.075, \dots, 0.975\}.
\]
Then $\mathrm{prob\_slope}_{20}$ is the ordinary-least-squares slope obtained by regressing $p_b(20)$ on $x_b$. Larger values mean that the backtrack profile rises later in the trace.

\paragraph{Late-half severe mean at $T \ge 50$.}
\[
\mathrm{prob\_late\_mean}_{50}
\deq
\frac{1}{10}\sum_{b=11}^{20} p_b(50).
\]
This measures whether severe backtracking persists in the second half of the trace.

\paragraph{Profile similarity at $T \ge 50$.}
For a trace from question $q$, let
\[
u_b^{(-q)} \deq \overline{p}^{(\mathrm{wrong},-q)}_b(50) - \overline{p}^{(\mathrm{correct},-q)}_b(50),
\]
where the class means are computed from other training questions only, excluding every trace from question $q$. Then
\[
\mathrm{profile\_similarity}_{50}^{(-q)}
\deq
\frac{\langle p(50), u^{(-q)} \rangle}{\|p(50)\|\,\|u^{(-q)}\|}.
\]
Large positive values mean that the trace resembles the held-out wrong-trace profile. For readability, we omit the superscript $(-q)$ elsewhere when the question holdout is understood.

The remaining four features are burden and burst summaries:
\[
r_{20}(D), \qquad S_{\ge 2}^{(20)}, \qquad \rho^{(40)}, \qquad m_{\max}^{(20)}.
\]
So the hybrid score is

\[
p_{\mathrm{wrong}}^{(\mathrm{hyb})}
\deq
\sigma\!\left(
\beta_0
 + \beta_1 z_{\mathrm{prob\_slope}_{20}}
 + \beta_2 z_{\mathrm{prob\_late\_mean}_{50}}
\right.
\]
\[
\left.
 + \beta_3 z_{\mathrm{profile\_similarity}_{50}^{(-q)}}
 + \beta_4 z_{r_{20}}
 + \beta_5 z_{S_{\ge 2}^{(20)}}
 + \beta_6 z_{\rho^{(40)}}
\right.
\]
\[
\left.
 + \beta_7 z_{m_{\max}^{(20)}}
\right).
\]

\subsection{Burst-Only Fallback}
\label{app:burst_only_fallback}

The burst-only filter discards normalized profile shape and keeps only burst organization:
\[
p_{\mathrm{wrong}}^{(\mathrm{burst})}
\deq
\sigma\!\left(
\beta_0
 + \beta_1 z_{S_{\ge 2}^{(20)}}
 + \beta_2 z_{\rho^{(40)}}
 + \beta_3 z_{m_{\max}^{(20)}}
\right).
\]
Its appeal is interpretability: if it rejects a trace, the reason is usually that the qualifying backtracks are compressed into larger and denser local clusters rather than appearing as isolated repairs.

\subsection{Burden-Only Auxiliary Filters}
\label{app:burden_only_aux}

These ablations test how far simple burden alone can go.

\paragraph{Rate-only auxiliary filter.}
Using only $r_{20}(D)$,
\[
p_{\mathrm{wrong}}^{(\mathrm{rate})}
\deq
\sigma(\beta_0 + \beta_1 z_{r_{20}}).
\]

\paragraph{Count-only auxiliary filter.}
Using only $N_{20}(D)$,
\[
p_{\mathrm{wrong}}^{(\mathrm{count})}
\deq
\sigma(\beta_0 + \beta_1 z_{N_{20}}).
\]

These two filters help separate the question ``is structure informative?'' from the question ``does a richer structural filter beat every simpler rule?''

\subsection{Online Prefix Filter}
\label{app:online_prefix_filter}

The prefix-causal filter is trained separately at each depth
\[
d \in \{2000, 5000, 8000, 12000\}
\]
and uses only events with start depth less than $d$.

Its feature set is
\[
r_{20}(d), \qquad S_{\ge 2}^{(20)}(d), \qquad \rho^{(40)}(d), \qquad m_{\max}^{(20)}(d),
\]
together with two timing features:
\[
h_{50}(d)
\deq
\min\{d_i : s_i \ge 50,\; d_i < d\},
\]
and
\[
\begin{aligned}
b_{40}(d)
\deq{} & \textnormal{start depth of the first} \\
& T \ge 40 \textnormal{ multi-burst before } d.
\end{aligned}
\]
The prefix score is therefore
\[
\begin{aligned}
p_{\mathrm{wrong}}^{(\mathrm{prefix})}
\deq
\sigma\!\big(&
\beta_0
 + \beta_1 z_{r_{20}(d)}
\\
&+ \beta_2 z_{S_{\ge 2}^{(20)}(d)}
 + \beta_3 z_{\rho^{(40)}(d)}
\\
&+ \beta_4 z_{m_{\max}^{(20)}(d)}
 + \beta_5 z_{h_{50}(d)}
\\
&+ \beta_6 z_{b_{40}(d)}
 \big).
\end{aligned}
\]

At $d=2000$, one extra rule is imposed:
\[
\textnormal{drop at 2k}
\iff
\bigl(p_{\mathrm{wrong}}^{(\mathrm{prefix})} \ge c\bigr)
\wedge
\bigl(N_{20}(2000) \ge 2\bigr).
\]
This 2k guard encodes the empirical point that one early correction should not be treated as a failure signature.
\section{Additional Filtering Results}
\label{app:additional_filtering_results}

This section expands the main-text filtering comparison. Completed-trace results compete against length-based baselines with full hindsight; prefix-causal results compete against a fixed stop-at-depth rule at each observed depth. All values use the question-controlled protocol defined in Appendix~\ref{app:filter_setup}.

\subsection{Hard-Cutoff Sweep}
\label{app:hard_cutoff_sweep}

The hard-cutoff sweep provides the baseline landscape that every completed-trace method must beat. It is also the cleanest way to show why a single global length rule is an incomplete account of reasoning quality. On this corpus, moderate cutoffs already perform strongly, but aggressive cutoffs rapidly become destructive.

\begin{table}[t]
\centering
\footnotesize
\setlength{\tabcolsep}{3.5pt}
\renewcommand{\arraystretch}{1.0}
\begin{tabular}{@{}lcccc@{}}
\toprule
\textbf{Limit} & \textbf{AIME24} & \textbf{AIME25} & \textbf{Word save} & \textbf{Drop} \\
\midrule
No limit     & 73.2\% & 64.5\% & 0.0\%  & 0.0\%  \\
$\le 12000$  & 80.6\% & 71.7\% & 32.2\% & 34.3\% \\
$\le 10000$  & 82.9\% & 68.0\% & 40.3\% & 44.3\% \\
$\le 8000$   & 71.9\% & 54.4\% & 48.4\% & 55.7\% \\
$\le 5000$   & 45.7\% & 35.9\% & 63.2\% & 78.3\% \\
$\le 2000$   & 10.0\% & 6.7\%  & 81.2\% & 99.8\% \\
\bottomrule
\end{tabular}
\caption{Selected hard-cutoff baselines. Word savings and drop rate are pooled across both years. Moderate cutoffs remain strong; harsh truncation is catastrophic.}
\label{tab:hard_cutoff_sweep}
\end{table}

The sweep makes the core asymmetry of the dataset very clear. A moderate cap around $10$k--$12$k words is already a strong completed-trace rule. The $12$k cutoff improves over the no-limit baseline on both years, and the $10$k cutoff is strongest on AIME2024. That leaves relatively little room for any more elaborate completed-trace filter to claim an unambiguous practical win.

At the same time, the lower rows show why a naive ``shorter is better'' interpretation is inadequate. Once the cap is pushed down to $5$k and especially $2$k, retained accuracy collapses. The same rule family therefore moves from useful to catastrophic over a fairly small range of depths. This is exactly the setting in which a trace-aware instability signal becomes worth studying: not because length is useless, but because length alone cannot distinguish a productive long trace from an unproductive one.

\subsection{Completed-Trace Structural Filters and Auxiliary Ablations}
\label{app:completed_aux_results}

The next comparison holds the completed-trace setting fixed and asks a narrower question: once the whole trajectory is visible, does backtracking structure provide a better keep/drop rule than length alone? On the present corpus, the answer is mixed and should be stated conservatively. The structural features are informative, but the strongest moderate hard cutoffs remain difficult to beat.

\begin{table*}[t]
\centering
\footnotesize
\setlength{\tabcolsep}{4pt}
\renewcommand{\arraystretch}{1.03}
\resizebox{0.9\textwidth}{!}{%
\begin{tabular}{@{}llcccc@{}}
\toprule
\textbf{Method family} & \textbf{Method} & \textbf{AIME2024 acc.} & \textbf{AIME2025 acc.} & \textbf{AIME2024 drop} & \textbf{AIME2025 drop} \\
\midrule
Hard cutoff baseline & Hard cutoff 12k & 80.6\% & 71.7\% & 29.9\% & 38.7\% \\
Hard cutoff baseline & Hard cutoff 10k & 82.9\% & 68.0\% & 38.6\% & 50.0\% \\
Structural & Completed hybrid & 77.4\% & 65.0\% & 44.8\% & 73.0\% \\
Structural & Burst-only fallback & 77.8\% & 68.0\% & 47.4\% & 55.6\% \\
Auxiliary burden-only & Raw-rate only & 78.3\% & 70.2\% & 54.9\% & 68.6\% \\
Auxiliary burden-only & Absolute-count only & 79.2\% & 69.4\% & 44.2\% & 52.6\% \\
\bottomrule
\end{tabular}%
}
\caption{Completed-trace filtering results and auxiliary ablations. The hybrid is the most faithful structural formulation, but not the strongest practical full-trace rule on this corpus.}
\label{tab:completed_aux_results}
\end{table*}

Relative to the strongest moderate hard cutoffs, the answer is no. Neither structural filter surpasses the $10$k/$12$k baseline family, and the completed hybrid in particular does not turn the paper's descriptive story into the highest retained-accuracy full-trace rule. This matters because it keeps the empirical claim narrow and credible: the structural analysis is informative, but it does not erase the fact that a simple moderate cap already works well on these traces.

Relative to harsh truncation, however, the picture is much more favorable. The burst-only fallback and even the hybrid are far less destructive than a $5$k cutoff while still removing a substantial share of traces --- better viewed as principled alternatives to overly aggressive global truncation than as universal replacements for the best moderate cutoff. The burden-only ablations sharpen this reading: simple burden alone already carries substantial signal, so the strongest conclusion is about what backtracking structure diagnoses, not about a guaranteed end-to-end win for the most elaborate composite rule.

\subsection{Prefix-Causal Filters and Matched Hard Cutoffs}
\label{app:prefix_aux_results}

The prefix-causal comparison is operationally different and more favorable to trace-aware filtering. Here the question is not whether a structural rule beats the best completed-trace baseline after seeing the whole sample. The question is whether, at a fixed observed depth, one can do better than a blunt stop-at-depth rule by looking at how revision unfolds inside the observed prefix.

\begin{table}[t]
\centering
\scriptsize
\setlength{\tabcolsep}{3pt}
\renewcommand{\arraystretch}{1.0}
\begin{tabular}{@{}lcccc@{}}
\toprule
\textbf{Depth} & \textbf{Hard cutoff} & \textbf{Online} & \textbf{Raw-rate} & \textbf{Online drop} \\
 & \textbf{(A24/A25)} & \textbf{(A24/A25)} & \textbf{(A24/A25)} & \textbf{(A24/A25)} \\
\midrule
$2000$  & 10.0/6.7  & 73.5/64.5 & 73.4/64.5 & 21.7/27.0 \\
$5000$  & 45.7/35.9 & 68.2/60.2 & 71.4/61.6 & 64.6/49.1  \\
$8000$  & 71.9/54.4 & 73.9/64.1 & 76.0/64.4 & 42.7/45.2 \\
$12000$ & 80.6/71.7 & 79.4/69.3 & 79.1/69.4 & 26.8/35.0 \\
\bottomrule
\end{tabular}
\caption{Prefix-causal filters versus matched hard cutoffs. Trace-aware filtering dominates harsh shallow truncation; at later depths, burden-only prefix ablations become competitive or slightly stronger.}
\label{tab:prefix_aux_results}
\end{table}

This is the clearest operational use-case for the backtracking signal. At $2$k words, the matched hard cutoff severely degrades performance, whereas the online prefix rule stays close to the no-filter baseline. That is exactly the regime in which the timing analysis matters: successful traces often make an early isolated repair, so a hard stop at $2$k confuses healthy correction with failure. The 2k guard is intended to prevent precisely that error.

The same qualitative point remains visible at $5$k. The online prefix rule is far above the fixed $5$k truncation on both years, which shows that shallow and intermediate-depth decisions benefit from looking at organization and timing rather than only whether a global budget has been crossed. In other words, when the cutoff is early enough to be dangerous, trace-aware filtering is where the practical value is most obvious.

At $8$k and $12$k, the picture becomes more competitive. The raw-rate prefix ablation is as good as or slightly stronger than the structural prefix filter, and the fixed $12$k cutoff remains very strong. So the supported claim is again narrower than a universal victory statement. Trace-aware early exit is dramatically better than harsh shallow truncation and remains competitive at intermediate depths, but later-depth completed-trace decisions can still be served well by a moderate hard cap.

Taken together, Section~\ref{app:additional_filtering_results} establishes an important asymmetry. The best completed-trace baseline in this corpus is simple, but the strongest shallow early-exit behavior requires more than length alone. That is the operational niche in which the backtracking signal earns its keep.

\subsection{Cross-Corpus Prefix-Causal Details}
\label{app:cross_corpus_prefix_details}

For completeness, Table~\ref{tab:cross_corpus_prefix} gives the same fixed-depth comparison for the additional model and domain pairs used in Section~\ref{sec:cross_corpus_replication}. This table is an operational supplement to the main cross-corpus signature rather than a matched benchmark, since the rows differ in sampling budget, trace length distribution, and judging protocol.

\begin{table*}[t]
\centering
\scriptsize
\setlength{\tabcolsep}{3pt}
\renewcommand{\arraystretch}{1.03}
\resizebox{0.9\textwidth}{!}{%
\begin{tabular}{@{}llcccc@{}}
\toprule
\textbf{Corpus} & \textbf{Model} &
\textbf{2k H/O/D} & \textbf{5k H/O/D} &
\textbf{8k H/O/D} & \textbf{12k H/O/D} \\
\midrule
AIME2024 & Qwen3-8B & 10.0/73.5/21.7 & 45.7/68.2/64.6 & 71.9/73.9/42.7 & 80.6/79.4/26.8 \\
AIME2024 & Phi4R & 43.9/72.1/1.7 & 68.4/71.4/17.0 & 74.8/73.0/21.3 & 80.1/78.3/14.7 \\
AIME2024 & Qwen3.5-9B & 0.0/94.7/29.3 & 13.3/93.3/4.0 & 53.3/93.8/14.7 & 80.0/92.0/3.3 \\
\midrule
AIME2025 & Qwen3-8B & 6.7/64.5/27.0 & 35.9/60.2/49.1 & 54.4/64.1/45.2 & 71.7/69.3/35.0 \\
AIME2025 & Phi4R & 26.2/67.8/5.3 & 63.8/65.6/11.0 & 72.9/74.4/27.3 & 75.2/73.0/13.3 \\
AIME2025 & Qwen3.5-9B & 0.0/83.3/19.3 & 20.0/82.5/10.7 & 46.7/81.2/10.7 & 72.5/84.7/16.7 \\
\midrule
GPQA-Diamond & Qwen3-8B & 40.6/60.5/24.2 & 60.1/61.1/5.5 & 60.9/60.8/1.3 & 60.6/60.6/0.0 \\
GPQA-Diamond & Phi4R & 47.9/65.7/9.5 & 61.4/68.6/23.2 & 67.4/67.8/16.2 & 67.8/67.6/6.6 \\
GPQA-Diamond & Qwen3.5-9B & 19.2/82.1/3.2 & 76.7/82.1/9.9 & 82.8/82.4/3.2 & 82.3/82.3/0.4 \\
\midrule
OmniMath & Qwen3-8B-thinking & 12.4/66.0/0.0 & 33.7/66.0/0.0 & 49.6/66.0/0.0 & 61.3/66.0/0.0 \\
OmniMath & Phi4R & 37.0/65.3/0.0 & 54.9/65.3/0.0 & 61.2/65.3/0.0 & 64.3/65.3/0.0 \\
OmniMath & Qwen3.5-9B & 7.0/70.5/0.0 & 20.3/70.5/0.0 & 37.4/70.5/0.0 & 57.8/70.5/0.0 \\
\bottomrule
\end{tabular}%
}
\caption{Cross-corpus analogue of the prefix-causal comparison in Table~\ref{tab:prefix_filtering_results}. Each cell reports hard cutoff accuracy / online prefix accuracy / online prefix drop rate. Horizontal rules separate datasets, not model families. The operational benefit is largest when fixed truncation is too shallow or mismatched to the corpus. The OmniMath rows show 0.0\% drop at every depth because the corpus has a single trace per question and the prefix filter is trained leave-one-question, so the online column reflects the no-filter baseline rather than an active filtering decision.}
\label{tab:cross_corpus_prefix}
\end{table*}

\section{Additional Evidence for Structural Signal}
\label{app:additional_evidence_structural_signal}

Because the completed hybrid is not the top full-trace decision rule, it is useful to separate two questions that are easy to conflate. The first is which final filter ranks highest on retained accuracy. The second is whether the structural variables highlighted by the descriptive analysis actually carry independent predictive information. This section addresses the second question. The answer is yes, and the evidence comes from two complementary directions: strong single-feature performance and formal incremental-value tests.

\subsection{Strongest Single Structural Signals}
\label{app:single_signal_results}

The single-feature view is valuable because it strips away interactions among variables and asks whether the main ingredients of the descriptive story are informative on their own. If the answer were no, then the hybrid's underperformance would be easy to interpret as evidence against the structural narrative itself. The results below point the other way. Each single-signal filter is the Appendix~\ref{app:filter_setup} template with one feature, $p_{\mathrm{wrong}} = \sigma(\beta_0 + \beta_1 z_x)$.

\begin{table*}[t]
\centering
\footnotesize
\setlength{\tabcolsep}{4pt}
\renewcommand{\arraystretch}{1.03}
\resizebox{0.9\textwidth}{!}{%
\begin{tabular}{@{}lclclc@{}}
\toprule
\textbf{Study} & \textbf{Threshold} & \textbf{Best signal on AIME2024} & \textbf{AIME2024 acc.} & \textbf{Best signal on AIME2025} & \textbf{AIME2025 acc.} \\
\midrule
Normalized trace rejection & $T \ge 20$ & Prob slope & 80.6\% & Prob slope & 69.8\% \\
Normalized trace rejection & $T \ge 50$ & Profile similarity & 82.2\% & Prob late mean & 68.2\% \\
Normalized trace rejection & $T \ge 60$ & Profile similarity & 82.0\% & Prob slope & 68.4\% \\
Burst-based rejection & $T \ge 20$ & Max burst size & 81.8\% & Max burst size & 70.4\% \\
Burst-based rejection & $T \ge 40$ & Compression ratio & 81.1\% & Compression ratio & 71.3\% \\
\bottomrule
\end{tabular}%
}
\caption{Strongest single structural signals from the profile-shape and burst studies. The ingredients of the hybrid remain informative even when the full composite rule does not win the final benchmark.}
\label{tab:single_signal_results}
\end{table*}

The main point is not that any one single feature should replace the final filters. The point is that the ingredients emphasized by the descriptive analysis are individually predictive. On the normalized-profile side, profile slope and profile similarity repeatedly emerge. On the burst side, maximum burst size and compression ratio recur. These are not just indirect proxies for total length; they encode whether instability appears late and whether it arrives in clustered form.

\subsection{Likelihood-Ratio Tests for Added Burst Signal}
\label{app:lr_tests}

The likelihood-ratio analysis asks a stricter question than the accuracy tables. Instead of comparing unrelated rules, it asks whether burst variables add information beyond simpler burden summaries once those burden summaries are already in the model. This is the right test for the narrower claim that burst organization contains genuine incremental signal.

\begin{table}[t]
\centering
\footnotesize
\setlength{\tabcolsep}{4pt}
\renewcommand{\arraystretch}{1.0}
\begin{tabular}{@{}cccc@{}}
\toprule
\textbf{Depth} & \textbf{Threshold} & \textbf{LR stat.} & \textbf{\(p\)-value} \\
\midrule
$2000$  & $T \ge 20$ & 16.133 & 0.001065 \\
$2000$  & $T \ge 40$ & 16.213 & 0.001025 \\
$5000$  & $T \ge 20$ & 38.407 & $2.318 \times 10^{-8}$ \\
$5000$  & $T \ge 40$ & 40.969 & $6.64 \times 10^{-9}$ \\
$8000$  & $T \ge 20$ & 11.038 & 0.01152 \\
$8000$  & $T \ge 40$ & 8.186 & 0.04233 \\
$12000$ & $T \ge 10$ & 19.567 & 0.0002087 \\
$12000$ & $T \ge 70$ & 14.785 & 0.00201 \\
\bottomrule
\end{tabular}
\caption{Selected pooled-corpus likelihood-ratio tests for adding burst features beyond simpler burden summaries. Burst organization contributes statistically significant signal across several evaluation depths and thresholds.}
\label{tab:lr_tests}
\end{table}

Across multiple depths and thresholds, the pooled-corpus likelihood-ratio statistics are significant, often strongly so. The especially large gains at $5$k are notable because that is also the regime where practical early-exit decisions are most consequential. The result does not prove that the burst-aware classifier must beat every simpler deployed rule on retained accuracy, but it does show that burst organization contributes measurable information beyond plain burden.

That distinction matters for the paper's framing. A reader can agree that the strongest completed-trace practical baseline is still a moderate hard cutoff while also recognizing that backtracking structure adds real explanatory and predictive value. The likelihood-ratio tests support exactly that middle position: not ``structure wins everything,'' but ``structure contains signal that length and burden alone do not exhaust.''

Taken together, the single-signal and likelihood-ratio results justify keeping the structural analysis central to the paper. Even when the best completed-trace rule is still a simple moderate cutoff, the profile-shape and burst results explain what that cutoff misses: successful traces often repair themselves early, whereas failing traces are more likely to enter later phases of clustered, repeated, and increasingly severe reversal. Level~2 is the strongest main-text regime, with Level~3 as supporting evidence; Level~1 is too loose to be very informative, and Level~4 is too sparse to carry the main story.

\section{Burst-Gap Robustness}
\label{app:burst_gap_robustness}

\begin{table*}[!t]
\centering
\scriptsize
\setlength{\tabcolsep}{3pt}
\renewcommand{\arraystretch}{1.0}
\resizebox{0.9\textwidth}{!}{%
\begin{tabular}{@{}llcccccc@{}}
\toprule
& & \multicolumn{3}{c}{\textbf{$T\!\ge\!20$ multi-bursts C/W}} & \multicolumn{3}{c}{\textbf{$T\!\ge\!50$ multi-bursts C/W}} \\
\cmidrule(lr){3-5} \cmidrule(lr){6-8}
\textbf{Corpus} & \textbf{Model} & \textbf{$g=250$} & \textbf{$g=500$} & \textbf{$g=1000$} & \textbf{$g=250$} & \textbf{$g=500$} & \textbf{$g=1000$} \\
\midrule
AIME2024 & Qwen3-8B          & $2.47 / 7.13$  & $1.90 / 4.78$ & $1.34 / 2.71$ & $1.47 / 5.09$  & $1.21 / 3.99$  & $0.97 / 2.69$ \\
AIME2024 & Phi4R             & $0.74 / 2.74$  & $0.62 / 2.35$ & $0.50 / 1.68$ & $0.66 / 2.29$  & $0.56 / 2.06$  & $0.46 / 1.61$ \\
AIME2024 & Qwen3.5-9B        & $7.79 / 13.50$ & $2.61 / 4.60$ & $1.29 / 1.60$ & $7.18 / 15.70$ & $4.67 / 10.10$ & $2.25 / 4.90$ \\
\midrule
AIME2025 & Qwen3-8B          & $2.87 / 6.30$  & $2.26 / 4.52$ & $1.52 / 2.78$ & $1.57 / 3.86$  & $1.42 / 3.36$  & $1.17 / 2.50$ \\
AIME2025 & Phi4R             & $0.65 / 3.15$  & $0.58 / 2.45$ & $0.50 / 1.50$ & $0.51 / 2.69$  & $0.48 / 2.21$  & $0.42 / 1.47$ \\
AIME2025 & Qwen3.5-9B        & $7.06 / 15.32$ & $2.40 / 4.96$ & $1.17 / 1.48$ & $5.77 / 18.40$ & $3.90 / 8.40$  & $2.05 / 3.08$ \\
\midrule
GPQA-Diamond & Qwen3-8B      & $1.07 / 1.54$  & $0.77 / 1.05$ & $0.64 / 0.83$ & $0.83 / 1.08$  & $0.66 / 0.89$  & $0.55 / 0.69$ \\
GPQA-Diamond & Phi4R         & $0.61 / 1.91$  & $0.47 / 1.37$ & $0.35 / 0.88$ & $0.55 / 1.77$  & $0.42 / 1.29$  & $0.31 / 0.84$ \\
GPQA-Diamond & Qwen3.5-9B    & $2.59 / 4.32$  & $1.63 / 2.32$ & $1.07 / 1.31$ & $1.95 / 3.61$  & $1.43 / 2.35$  & $1.01 / 1.40$ \\
\midrule
OmniMath & Qwen3-8B-thinking & $2.28 / 6.08$  & $1.61 / 3.60$ & $1.06 / 1.93$ & $1.63 / 4.90$  & $1.31 / 3.46$  & $0.95 / 2.14$ \\
OmniMath & Phi4R             & $0.64 / 2.09$  & $0.52 / 1.64$ & $0.39 / 1.16$ & $0.53 / 1.69$  & $0.46 / 1.43$  & $0.36 / 1.08$ \\
OmniMath & Qwen3.5-9B        & $5.84 / 12.40$ & $3.00 / 5.29$ & $1.40 / 2.00$ & $4.66 / 11.94$ & $3.00 / 6.42$  & $1.64 / 2.73$ \\
\bottomrule
\end{tabular}%
}
\caption{Class-mean multi-burst count $K_{\ge 2}$ at thresholds $T\!\ge\!20$ (Level 2) and $T\!\ge\!50$ (Level 3) across all 12 model $\times$ dataset pairs from Section~\ref{sec:cross_corpus_replication}, at three burst-gap settings ($g \in \{250, 500, 1000\}$ words). Each cell reports correct/wrong class means; $g=500$ is the value used in the main text. The wrong $>$ correct ordering holds in all 72 (corpus, threshold, $g$) cells.}
\label{tab:burst_gap_robustness}
\end{table*}

Two questions about the burst construction in Section~\ref{sec:setup} naturally arise: does the wrong $>$ correct ordering survive a different choice of $g$, and does it generalize beyond the primary Qwen3-8B AIME2024 corpus? Table~\ref{tab:burst_gap_robustness} reports multi-burst counts at $g \in \{250, 500, 1000\}$ for every model $\times$ dataset pair examined in Section~\ref{sec:cross_corpus_replication}, at both the central Level-2 ($T\!\ge\!20$) and Level-3 ($T\!\ge\!50$) thresholds.

The wrong $>$ correct ordering holds in every cell: 72 of 72 (corpus, threshold, $g$) combinations agree on direction. The size of the gap shrinks as $g$ grows, which is mechanical --- larger gaps absorb more events into single long bursts, deflating the per-class multi-burst count for both classes while preserving the directional asymmetry; smaller gaps enforce stricter clustering, so each multi-burst becomes more selective and the heavier wrong-trace tail of locally repeated events separates more visibly. The same pattern shows up in the auxiliary burst statistics (event share in multi-bursts $S_{\ge 2}$ and maximum burst size $m_{\max}$), omitted from the table for space. The descriptive claim of the work is therefore not gap-tuned and not specific to the primary corpus: the choice $g=500$ used in the main text is a reasonable middle setting that balances burst granularity against over-aggregation, and the qualitative wrong $>$ correct asymmetry is recovered across every model and domain pair we tested.

\end{document}